# OntoAna: Domain Ontology for Human Anatomy

Archana Vashisth, Iti Mathur, Nisheeth Joshi
Department of Computer Science
Apaji Institute
Banasthali Univeristy
Rajasthan, India
archana.sharma3010@gmail.com, mathur_iti@rediffmail.com, nisheeth.joshi@rediffmail.com

*Abstract-* **Today, we can find many search engines which provide us with information which is more operational in nature. None of the search engines provide domain specific information. This becomes very troublesome to a novice user who wishes to have information in a particular domain. In this paper, we have developed an ontology which can be used by a domain specific search engine. We have developed an ontology on human anatomy, which captures information regarding cardiovascular system, digestive system, skeleton and nervous system. This information can be used by people working in medical and health care domain.**

## I. INTRODUCTION

These days, ontologies are being considered as the most important knowledge representation technique because they can semantically capture information. That's why they are being termed as the backbone of knowledge systems. A large number of applications are using knowledge databases for searching information in a particular domain, on a particular topic [1].

The term ontology has been taken from the Greek words *ontologia* which means talking about beings. It is just a term in philosophy & which means "theory of existence". This term was first coined by Plato in 3rd century B.C. Later Aristotle (his student) shaped the logical background of ontologies and introduced notions like category, subsumption, genus and subspecies. Aristotle's ideas represent the conceptual foundations of object oriented systems of today. Moreover, he developed a number of inference rules, called syllogisms, which are being used by a number of logic-based reasoning systems.

In today's era of computer science, one does not talk about the ontology as the science of existence anymore. However we consider ontology as a formal specification of concepts and thus creating a collection of such concept hierarchies, as described by Gruber [2]. Thus, ontology is knowledge representation technique which represents conceptual information in a particular domain, describing it in a declarative manner and in turn clearly separating it from procedural aspects. By declarative we mean that an ontology should be formal (should be machine readable), explicit (all concepts and constraints are explicitly defines), shared (should capture consensual knowledge accepted by stakeholders involved) and conceptual (describe abstract model phenomena in a real world & identify relevant concept of those phenomena). Thus, Ontology defines the formal, explicit, and shared representations of concepts, objects, and property which define relations between them.

According to Guranio [3], "An Ontology is generally regarded as a designed artifact consisting of a specific shared vocabulary used to describe entities in some domain of interest, as well as a set of assumptions about the intended meaning of the terms in the vocabulary."

## II. ONTOLOGY DEVELOPEMNT TOOLS

Some of the popular ontology development tools are Protégé, OntoEdit, WebODE, and Ontolingua. Ontologies can be created through any of these ontology editors. Using these tools we can create, browse, codify, and modify ontologies and in turn support ontology development and maintenance task. These editors vary in usability, modeling, scalability etc.

Among these Protégé, is the most popular editor, as it is being used by majority of developers throughout the world. It is an open source, free tool developed by Stanford University. Java developers can use the ontologies developed through this tool. Moreover, they can directly use Protégé APIs can call the environment directly.

## III. LITRATURE REVIEW

Development of computational ontologies was first introduced by Hearst who started working in the areas of concept annotation. Even today, his seminal work on lexico-syntactic patterns[4] is very relevant for annotation based knowledge applications. Hearst work has been refined and reused by several researchers who have applied different approaches. For example Poesio et al extended Hearst patterns for anaphora resolution [5] and using machine learning approaches in identifying patterns[6]. Eezioni [7] and Markert [8] have shown how these ontologies can be used on the Internet by using search engine APIs. Some researchers have also applied Lexico-syntactic patterns in the identification of other lexical relations like part-of relations [9] and causal relations [10][11] Cederberg and Widdows have shown that the precision of Hearst patterns can be improved by filtering the results of pattern matching using Latent Semantic Analysis [12]. Morin and Jacquemin[13] and Ravichandran and Hovy [14] has addressed the automatic generation of patterns via a similarity based approach where patterns are represented as vectors and if are found to be somewhat similar, are grouped together. This approach is more generalized then what was proposed by Hearst.

In literature, we can find systems which have been developed using these techniques like OntoLT [15] which is an ontology learning plugin  for Protégé Ontology editor. This system annotates parts of speech chunks, and grammatical relations using a parser. OntoLearn [16] is another system where terms are extracted for a certain domain from a domain-specific textual corpus. This tool has become one of the most important tools in automatic creation of Ontologies through text.

## IV. PROPOSED WORK

In order to develop ontology for human anatomy, we divided our work into four major sub area – namely cardiovascular system, digestive system, skeleton structure & nervous system. For each sub-system we defined classes, subclasses, disjoint classes, and equivalence classes.

For skeleton structure sub-system, we defined subclasses as front view, back view, human skull(side view), skull(above view), skull(inside view), human joint & mechanical equivalent, knee(skeleton view), hip(skeleton view), elbow(skeleton view), elbow(ligament view), shoulder(skeleton view), hand(bones ligament muscles) & teeth etc.

For nervous sub-system, we defined subclasses like human facial nervous, nerve cell, tongue(taste areas), ear(cut view), skin (cut view), eye’s rode & cones(cut view), brain(cut & surface view) etc.

For digestive sub-system, we included subclasses like mouth, human spleen, human stomach and gallbladder, liver, human intestine & human throat etc.

For cardiovascular sub-system we included subclasses like human heart picture (surface view), heart (cut view), kidney blood filtering, Lungs (cut view), coronary bypass (heart & leg view) & heat pacemaker etc.

### 3.1 Methodology

In this section we have shown a detailed description of phases for development of our ontology on human anatomy.

3.1.1 Specification

In this phase we collected information on our identified sub-systems from books, journals and web search engines. We did a comprehensive study of all the finer points. When we had certain doubts, we contacted different experts in the respected domains. Once all the information was collected, we ascertained the domain and scope of our ontology.

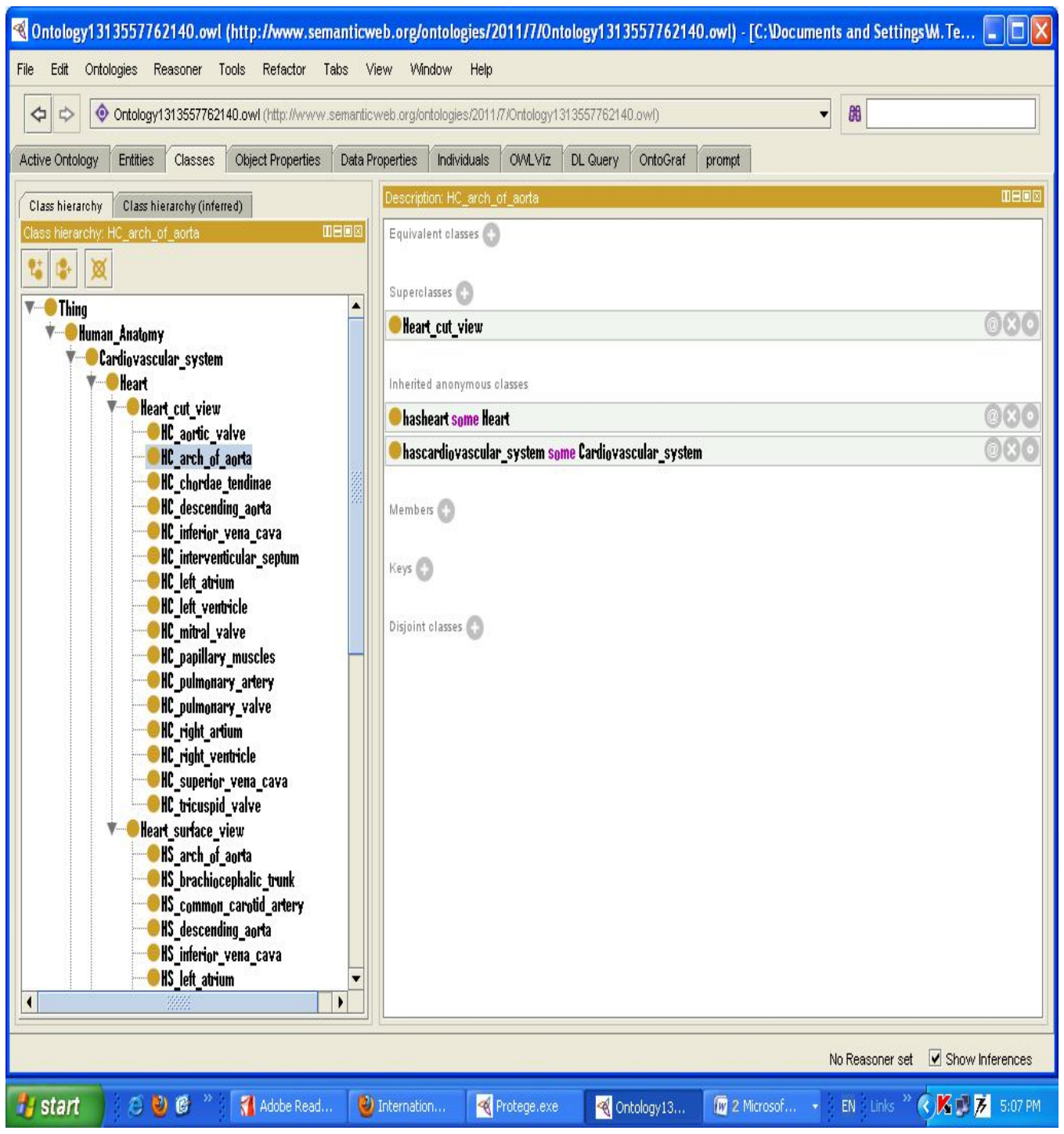

Fig.1 Structure of Human Anatomy Classes

3.1.2 Conceptualization

Once the specifications of domain and scope of our ontology was completed, we started the construction of concepts (classes) for our ontology sub-systems. In this we created the human anatomy as the super class and all the sub-systems as its sub classes. At times we came across classes with same properties and relations which were available in different sub-systems. These classes were marked at equivalence classes. Moreover, we also came across some of the classes which had some of the same properties across sub-systems. These classes were marked as disjoint classes. Figure 1, illustrates these concepts.

3.1.3 Relation

Once we defined concepts, classes and object were defined; we started the construction of relations between them. Here we used term property for defining the relation. Here we defined the internal structures of the classes through various properties like term property which defined the relations, object property which defined the relation between two relations.

The object property represented binary relations between individuals, for example, Hascardio_vascular_system was linked with cardio_vascular system in the body class. Figure 2, shows this property as defined in our ontology.

Various values of the classes were also defined through data type property, where we defined the type of value a class can accept. We also used the annotation property for annotating or describing classes and sub-classes. This acted like a label which can help others understand the concept and utility of the classes and sub-classes.

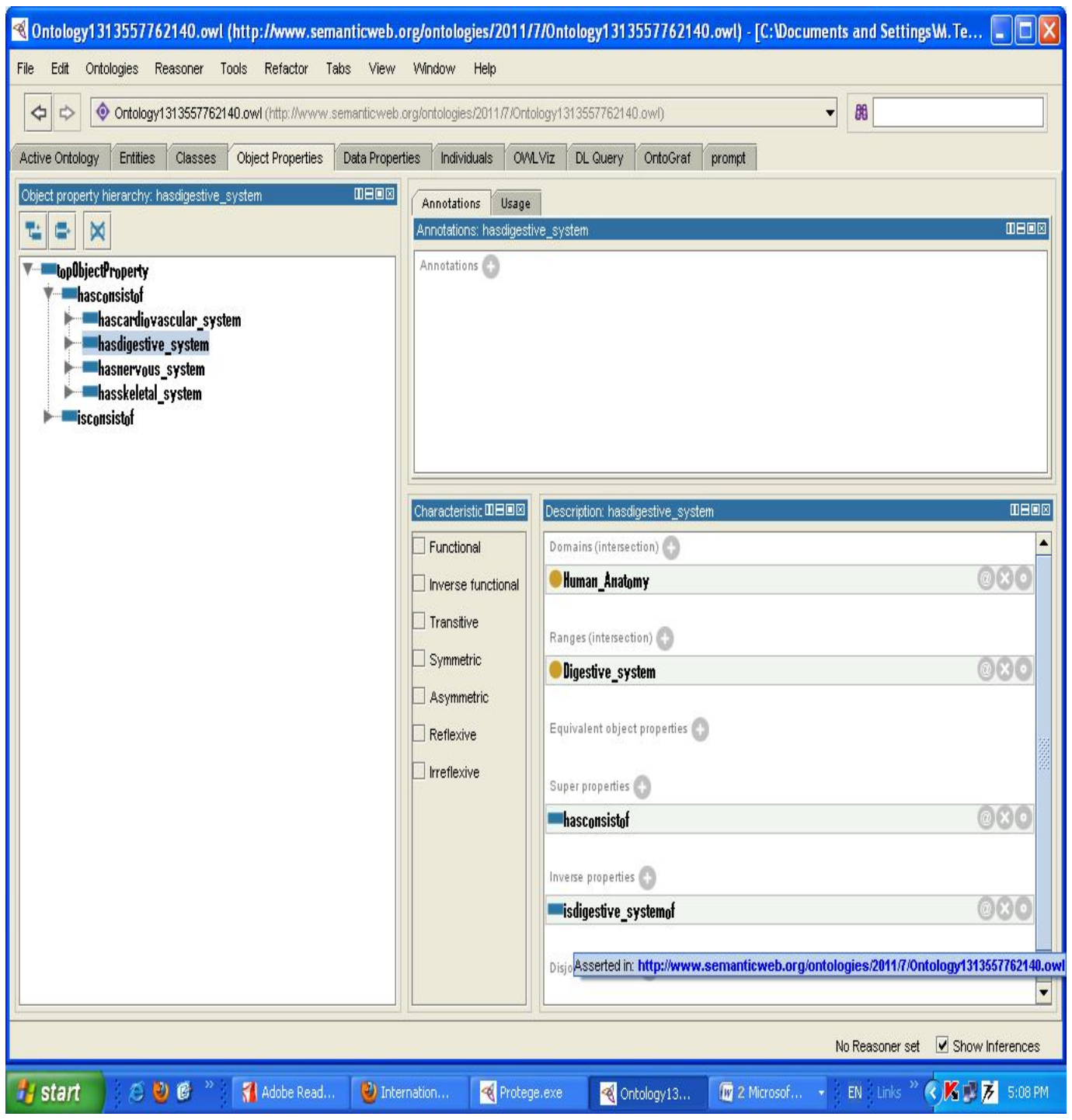

Fig. 2 Object property of system

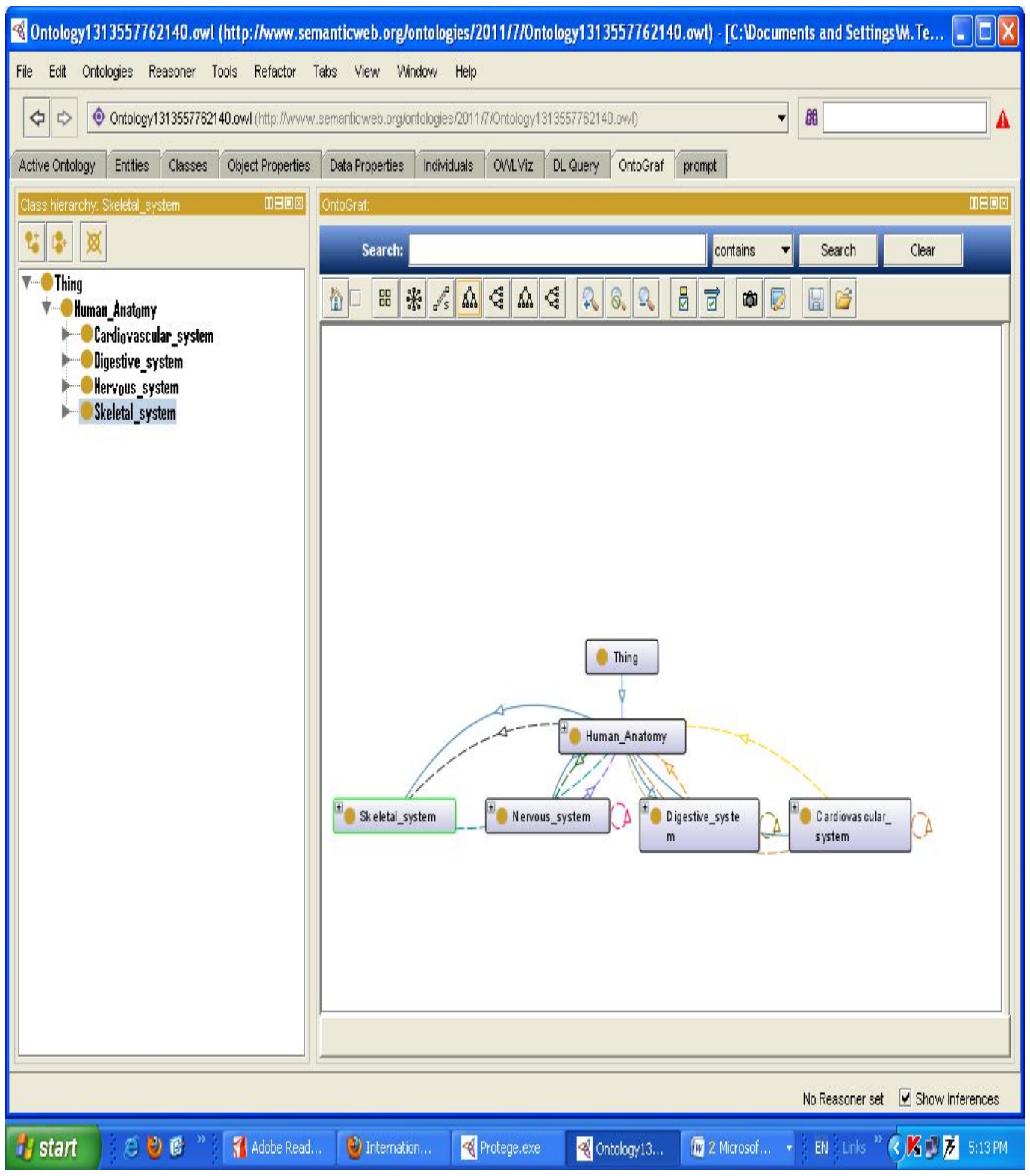

Fig.3 System structure

### 3.1.4 Structural view of the system

In this phase, we defined the graphical representation of the our ontology. As we have used Protégé 4.1 for the development of our ontology, we used Onto Graf, which is a built-in visualization tool for Protégé 4.1. Lower versions of Protégé do not have this feature, for this one needs to download GraphViz from sourceforge.net and link it with OntoViz visualization tool of Protégé. Figure 3 shows the structural view of our four major sub-systems.

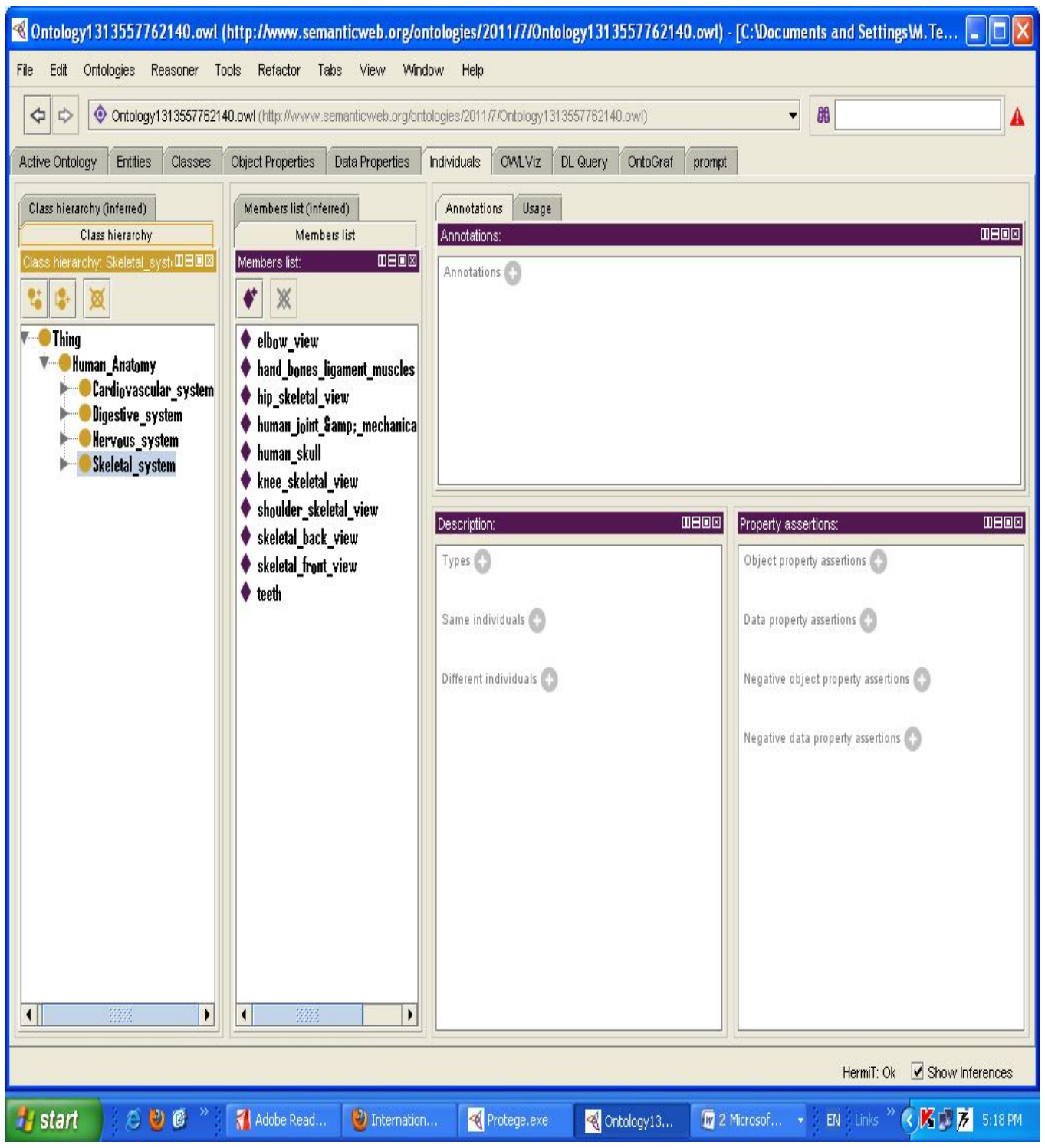

Fig. 4 Create Instances

### 3.1.5 Creation of instances

In this phase, we created the individual instances of the classes. In order to differentiate classes according to their components, we used individual properties and relations for each instance. Figure 4 shows the instance diagrams.

While clicking on the active ontology tab, we can view the summary of our ontology which displayed the no. of super class, subclasses, disjoints classes of our ontology being used. Figure 5 shows the diagram of our active ontology.

## V. EVALUATION

We needed to verify whether we had created the correct ontology or not. For this we used the DL Query feature of the of the Protégé, through which we checked each relation, property and in turn the classes. In this, we configure the reasoner for description logics, determining classes, instances, domain & range. Then, we used the query tab of the Protégé. Here we wrote the class name and when the query was executed, protégé retrieved the query in terms of classes, individuals, super class, domain and range. As all the desired relations and properties were retrieved, we were assured that our ontology has been created without any conflicts and errors. Figure 6 shows the diagram of query execution.

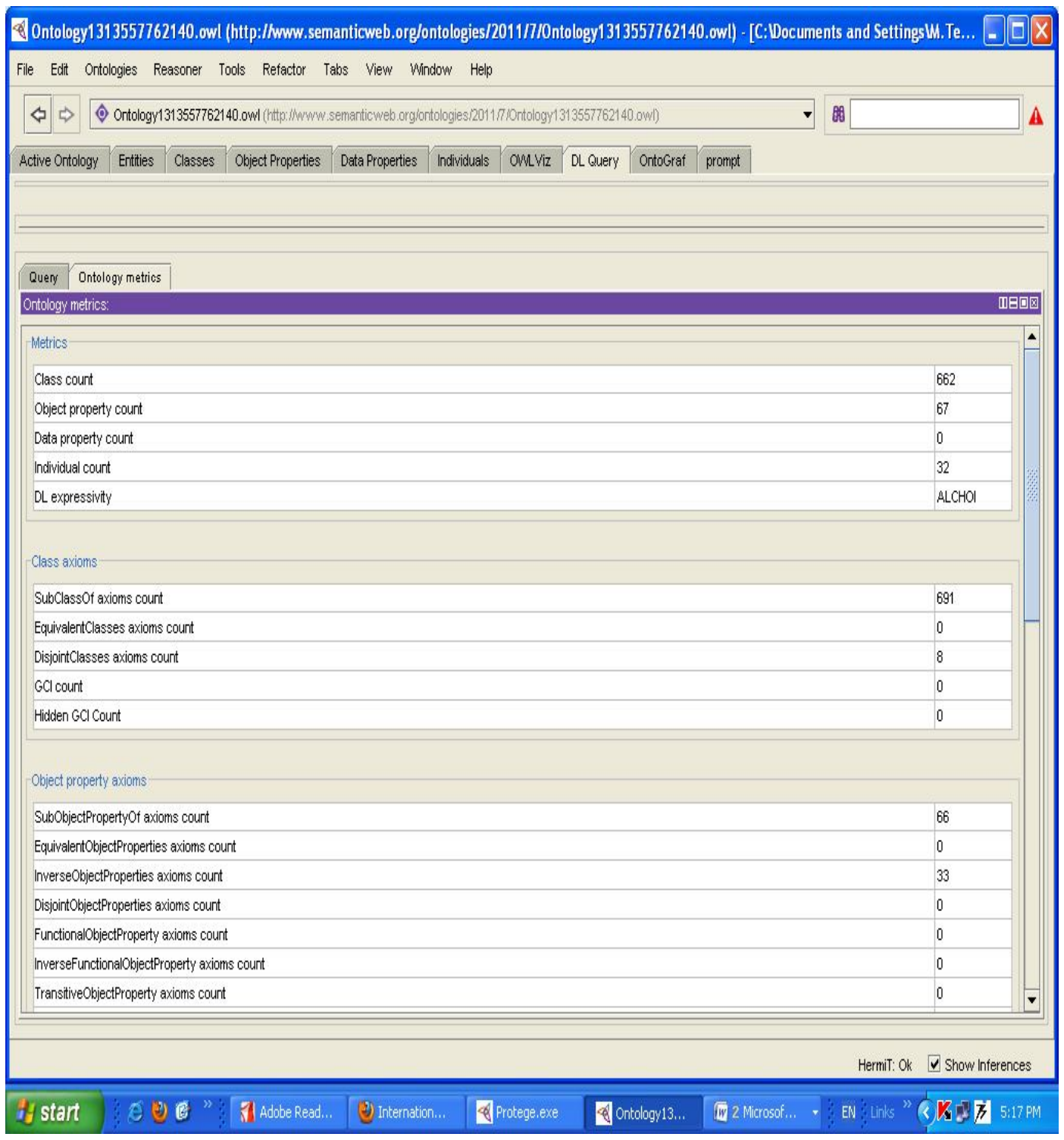

Fig.5 Active Ontology

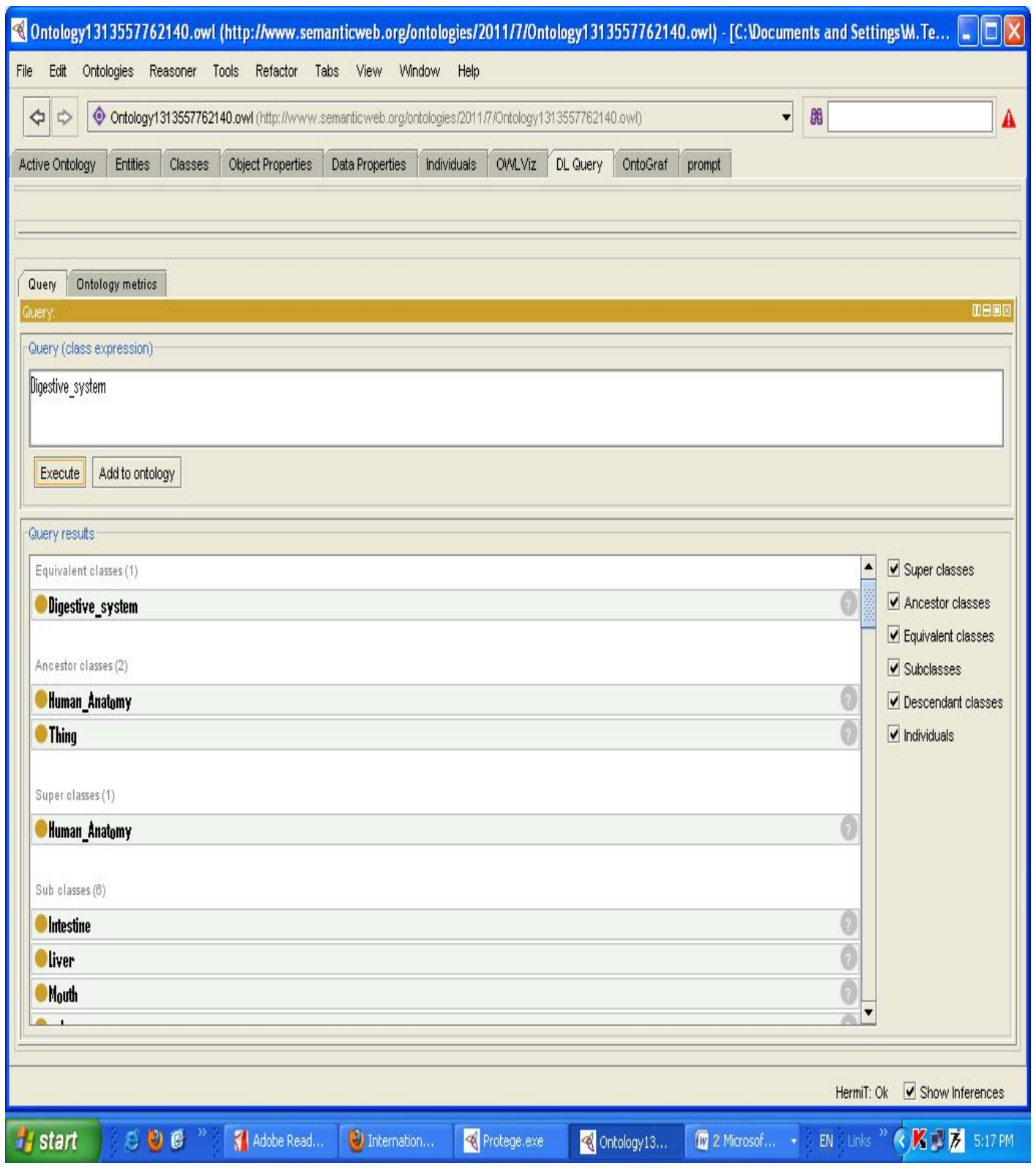

Fig 6 Query Execution

## VI. CONCLUSION

We have the shown the creation of an ontology for human anatomy. In this ontology we mainly used four sub-systems – cardiovascular system, digestive system, skeleton structure & nervous system. We have provided a detailed description of the creation of this ontology by explaining various phases of development. We have also verified the ontology by executing each class and their properties. In future wish to add more part of human body & reuse this ontology.